\documentclass[numbered]{trbunofficial}

\usepackage{graphicx}
\usepackage{booktabs}
\usepackage{array}
\usepackage{amsmath,bm}
\usepackage{makecell}
\usepackage{multirow}
\usepackage{xstring}

\graphicspath{{./}}

\nolinenumbers
\newcommand{\captitle}[1]{{\bfseries\boldmath #1}\ \normalfont\unboldmath\mdseries}

\begin{document}

\makeatletter
\long\def\@ifpackageloaded#1#2#3{#2}
\makeatother

\title{Pre-Lane-change Signal in Transitional Autonomous Vehicles: Results from Controlled Experiments}

\TRBauthor{Zeyu Mu}
{Department of Systems and Information Engineering, University of Virginia}
{dwe4dt@virginia.edu}
[Charlottesville, VA 22903]

\TRBauthor*{Danjue Chen}
{Department of Civil, Construction, and Environmental Engineering, North Carolina State University}
{dchen33@ncsu.edu}
[Raleigh, NC 27695, USA]

\TRBauthor{Abhinav Sharma}
{Department of Civil, Construction, and Environmental Engineering, North Carolina State University}
{asharm63@ncsu.edu}
[Raleigh, NC 27695, USA]

\TRBauthor{George F. List}
{Department of Civil, Construction, and Environmental Engineering, North Carolina State University}
{gflist@ncsu.edu}
[Raleigh, NC 27695, USA]

\AuthorHeaders{Mu, Chen, Sharma, List}

\maketitle

\section{Abstract}

\noindent\textbf{Objectives:} This paper investigates how a production transitional autonomous vehicle (tAV) develops and executes mandatory lane-change decisions. The objectives are to examine whether information about the eventual target gap is observable before lateral movement begins and how the tAV progresses longitudinally from that pre-lane-change state to lane-change start.

\hfill\break
\noindent\textbf{Methods:} The analysis uses 150 controlled mandatory lane changes from the NC-tALC experiments. Signal time (SigT) is defined as an operational pre-lane-change-start reference point. A Firth logistic regression predicts whether the tAV eventually merges in front of or behind its nearest target-lane vehicle using relative position and relative speed at SigT. Longitudinal progression from SigT to lane-change start is then examined separately for in-position and repositioning cases.

\hfill\break
\noindent\textbf{Findings:} The traffic state at SigT contains substantial information about the eventual target-gap choice and provides meaningful lead time before lateral movement begins. The proposed formulation predicts whether the tAV remains with the gap it occupies at SigT or repositions to a neighboring gap by moving forward or dropping back, including cases with longitudinal overlap and ambiguous current-gap geometry. The model attains an average five-fold cross-validated accuracy of 0.89. The analysis also provides preliminary evidence that in-position and repositioning cases follow different longitudinal pathways from SigT to lane-change start.

\hfill\break
\noindent\textbf{Novelty:} The findings support a two-stage conjecture for the observable lane-change process: longitudinal preparation from SigT to lane-change start, followed by lateral maneuver execution. The front/behind formulation applies to in-position, repositioning, and longitudinally overlapping cases.

\hfill\break
\noindent\textbf{Practical Applications:} The findings can support lane-change models that distinguish target-gap choice from lateral-onset timing and represent longitudinal preparation before lateral movement begins.

\medskip
\noindent\textit{Keywords:} transitional autonomous vehicle; lane changing; gap choice; longitudinal preparation; lane-change decision; controlled experiment.

\newpage

\section{Introduction}\label{sec:intro}
Recently, vehicles equipped with advanced driving-assistance capabilities have been operating on roads \cite{Ammourah2024IntroductionExtraction,TeslaInc.2025ModelSupervised,GeneralMotors2025HowCruise,FordMotorCompany2025BlueCruise,Mattas2025SafetyExperiments} and are capable of independently performing lane-changing (LC), including both LC decision making and lateral maneuver execution, under driver supervision.  This includes determining whether an LC is needed and executing the maneuver to move from the vehicle’s original lane to the target lane. We refer to vehicles with such capabilities as the transitional autonomous vehicles (tAVs), as they are distinct from vehicles at other automation levels.  Notably, this capability differs from driver-initiated assisted LC systems \cite{Mattas2025SafetyExperiments}, in which the human makes the LC decision and the automation primarily carries out the maneuver. Because both target-gap selection and lateral maneuvering are governed by the controller, understanding how a tAV develops its LC decision is particularly important.

The LC behavior of production tAVs has not yet been empirically examined. Fortunately, two controlled tAV experiments have been conducted, producing the NC-tALC datasets \cite{sharma2026controlled}. In these experiments, a tAV repeatedly performed mandatory LCs on a public road while the initial traffic conditions, including relative spacing and relative speed, were systematically varied. The resulting high-resolution trajectories provide a unique opportunity to study the tAV’s LC behaviors.

In this study, we are interested in how the tAV develops an LC decision and executes it in the real-world operation based on the NC-tAV dataset. 

The majority of the literature on lane-changing (LC) focuses on the LC decision problem and often formulates it as a gap-acceptance problem, asking whether a vehicle will accept or reject the gap it currently faces. There are different modeling frameworks.  The critical-gap models are among the classic and most broadly used frameworks, which compare whether the candidate gap exceeds the critical gap.  The critical gap may use fixed thresholds or probabilistic formulations \cite{ToledoKatz2009,Kim2008,Marczak2013MergingDevelopment}.  Other popular frameworks include utility-based models \cite{TOLEDO2007,Hess2020} and game-theoretic models \cite{KITA1999,TALEBPOUR2015,ALI2019,ALI2021} that consider both the lane-changer and interactive players in the target lane.
A smaller group of studies extends the classic formulation and reformulates LC as a multi-gap selection problem, e.g., \cite{Choudhury2009,Chu2017,Wan2017}. For example, \cite{Choudhury2009} developed a model that integrated gap choice and acceleration, which considered whether a vehicle accepts the current adjacent gap. If not, the vehicle further evaluates the options of moving forward or backward toward neighboring gaps.
Operationally, most LC decision models are tested against the observed outcome of whether the vehicle changes lanes. Specifically, in earlier work, this effectively treats two components, a positive LC decision (“Yes” for LC) and the lateral maneuver execution (moving from the original lane to the target lane), as occurring within the same observation interval, e.g., \cite{Choudhury2009,ToledoKatz2009}. This is likely because the available data had coarse temporal resolution. More recent studies recognized that the lateral maneuver execution is a substantial process that can last several seconds and thus focused on modeling the LC decision timing \cite{PENG2015,Bakhit2017,Wirthmuller2021,Mozaffari2022}. For example, \citet{PENG2015}, \citet{Bakhit2017},
\citet{Wirthmuller2021}, and \citet{Mozaffari2022} proposed models to predict the LC decision timing for HDVs, connected vehicles, and automated vehicles.  In these studies, the start of lateral movement (LC start, or LCS), is commonly used to label the decision, with “Yes” assigned at LCS and “No” assigned to earlier time steps. However, as pointed by \citet{ALI2023-2issues}, the actual LC decision occurred earlier than the onset of lateral movement and such treatment (equating LC decision point with lateral movement onset) is a compromise as the true LC decision is not observable in trajectory data.

It is worth noting that some studies, such as \cite{Choudhury2009}, conceptually separate gap selection from maneuver execution. However, gap selection was treated as a latent component that affects observable outcomes (such as whether an LC occurs and the vehicle’s acceleration) but the gap-selection component was not directly observed or tested.
Overall, the literature generally does not separate the gap decision component from the lateral maneuver execution and operationally assumes that the LC decision coincides with lateral-movement onset.

This paper investigates the mandatory LC process of a tAV using the NC-tAV datasets. We have two major results.  Firstly, a pre-LC state, the traffic state at SigT (signal time), contains substantial information about the tAV’s eventual target-gap choice, providing meaningful lead time before lateral movement begins. The proposed formulation predicts whether the tAV remains with the gap it faces at SigT or repositions to a neighboring gap by moving up or dropping back. It also applies to cases when the tAV longitudinally overlaps a target-lane vehicle and the current-gap geometry is ambiguous. This finding supports a two-stage conjecture for the observable LC process: longitudinal preparation from SigT to lateral-movement onset at LCS, during which lateral movement remains negligible, followed by lateral maneuver execution during which the tAV moves from its original lane to the target lane. We further conjecture that a target-gap decision-making process may precede longitudinal preparation, although this study is not able to further evaluate that due to data limitations.  Secondly, the study provides preliminary evidence that in-position cases, in which the tAV remains in the gap it occupies at SigT, and repositioning cases follow different longitudinal pathways from SigT to LCS, motivating further research on the longitudinal preparation process.

The remainder of the paper is organized as follows. Section 2 describes the dataset and key LC events. Section 3 presents the motivating observations and two-stage conjecture. Section 4 examines target-gap information at SigT, and Section 5 analyzes longitudinal preparation from SigT to LCS. The final section summarizes the findings and discusses implications and future research.

\section{Data}\label{sec:data}

\begin{figure}[!htbp]
    \centering\includegraphics[width=0.8\linewidth]{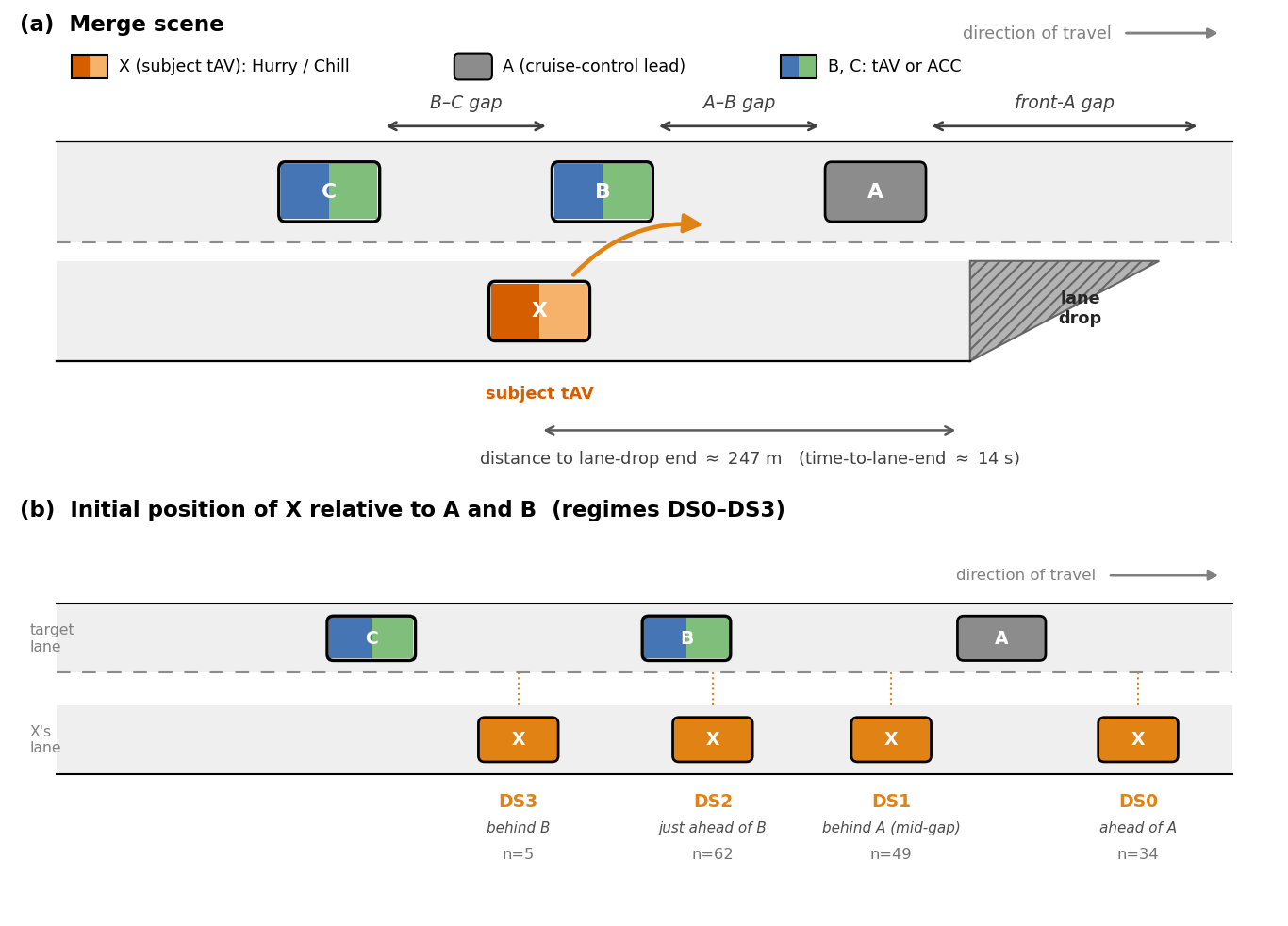}
    \caption{\captitle{The Mandatory-Merge Scenario.} (a) $X$ (a tAV) must merge from an ending lane into the adjacent target lane before a downstream lane drop, choosing among the \emph{front-A}, \emph{A-B}, and \emph{B-C} gaps; the lead A cruises, while followers B and C are tAV (sharing one Hurry/Chill mode, two-tone boxes) or ACC, and $X$ carries its own Hurry/Chill mode. (b) The four regimes DS0 to DS3 fix $X$'s initial longitudinal position relative to A and B (case counts 34/49/62/5).}
    \label{fig:data}
\end{figure}

The analysis uses 150 mandatory lane-change trajectories from the North Carolina Transitional Autonomous Vehicle Lane-Changing (NC-tALC) controlled field experiments. In each trial, the subject vehicle X, operating as a transitional automated vehicle (tAV), traveled in an ending lane and was required to merge into the adjacent target lane. Vehicles A, B, and C occupied the target lane, with A serving as the target-lane leader and Band C serving as potential followers. Vehicle A operated under cruise control, whereas B and C operated in either tAV or adaptive-cruise-control (ACC) mode. The four-vehicle arrangement created three candidate target gaps for X: ahead of A (front-A), between A and B (A-B), and between B and C (B-C), as illustrated in Figure \ref{fig:data}. The gaps between A, B, and C were determined solely by the control algorithm of B and C. We did not intervene. In all trials, X completed the merge successfully, with B and C adjusting their trajectories as needed to accommodate X.

The experiments varied the initial longitudinal position and relative speed of X with respect to the target-lane vehicles at automation activation. Before activation, X was manually positioned to establish the prescribed initial condition. At automation activation, denoted ACT, the automated driving system assumed control and independently conducted the mandatory lane change under continuous driver supervision. The analysis combines 78 cases with tAV followers and 72 cases with ACC followers, yielding 150 lane-change cases in total. 

Each vehicle was instrumented with an RTK-GNSS/INS unit that recorded synchronized position, speed, heading, and acceleration at 20 Hz. The trajectories were transformed into a common road-based coordinate system, with longitudinal position measured along the target-lane centerline and lateral position measured relative to that reference. These processed trajectories provide the vehicle states used in the subsequent analyses of signal time, target-gap choice, repositioning, and lane-change initiation. 

\begin{figure}[!htbp]
    \centering\includegraphics[width=0.8\linewidth]{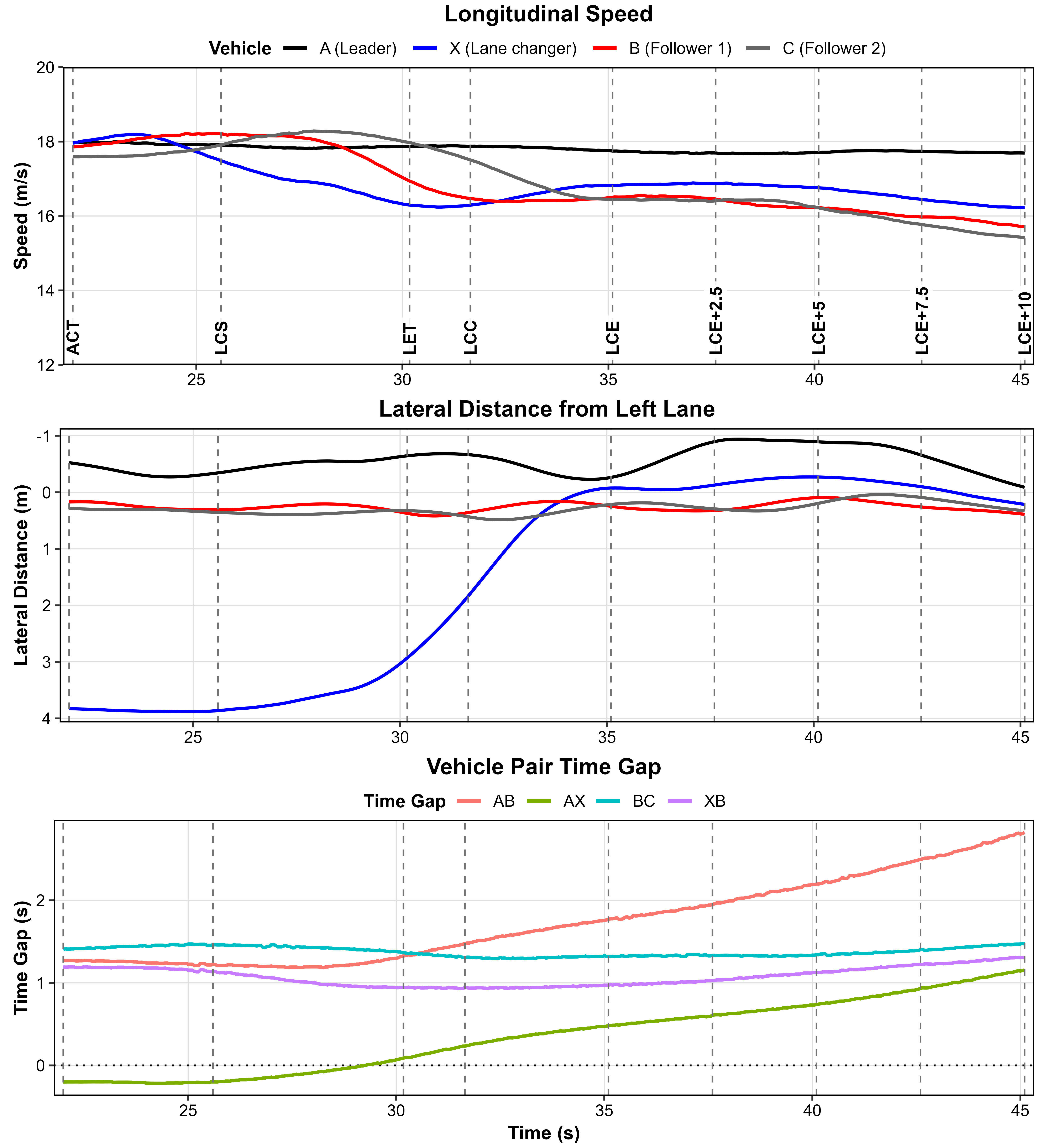}
    \caption{\captitle{Example trajectories of the four experimental vehicles during a mandatory lane change.} The panels show the longitudinal speed, lateral distance from the left lane, and vehicle-pair time gaps, respectively. The vertical dashed lines indicate the key lane-change timestamps: automation activation (ACT), lane-change start (LCS), left-edge touching (LET), lane-change crossing (LCC), lane-change end (LCE), and the post-lane-change timestamps at 2.5, 5, 7.5, and 10~s after LCE. \cite{sharma2026controlled}}
    \label{fig:key-timestamp-example}
\end{figure}

Each lane-change case is characterized by a sequence of key events. ACT marks automation activation. Signal time, denoted SigT, is an operationally identified pre-LCS reference point: for cases with a delayed sustained kinematic switch after ACT, SigT is defined as the observed switch time; otherwise, SigT is set equal to ACT. SigT is treated as an observable signal associated with target-gap information and longitudinal preparation, not as the confirmed instant at which the tAV internally selected its target gap. Lane-change start (LCS) marks the onset of lateral movement, left-edge touching (LET) marks the first physical entry of the vehicle into the target lane, lane-change crossing (LCC) marks the crossing of the vehicle center over the lane boundary, and lane-change end (LCE) marks completion of the maneuver and lateral stabilization in the target lane. We have introduced the definitions of ACT, LCS, LET, LCC, and LCE in our previous study, the NC-tALC Overview paper \cite{sharma2026controlled}; SigT is introduced specifically for the present analysis. Figure \ref{fig:key-timestamp-example} illustrates these key timestamps, including ACT, LCS, LET, LCC, and LCE, as presented in the overview paper. 

\section{OBSERVATIONS AND THE TWO-STAGE CONJECTURE }\label{sec:motivation}

Auto mode start (ACT) marks the instant the tAV assumes control of both decision making and execution, with the human monitoring but not maneuvering. In every case, the lane-change blinker activates immediately after ACT, signaling that X intends to change lanes. What follows differs across cases in the longitudinal trajectory, even though X’s lateral position remains approximately unchanged until LCS.

In a subset of cases, X first continues the longitudinal kinematics present at ACT and then, before LCS, exhibits a sustained kinematic switch, characterized by a reversal in the sign of longitudinal jerk, such as a transition from acceleration to deceleration or the reverse. The longitudinal adjustment following the switch continues toward LCS, while the gap ultimately used by X remains generally unchanged. In 23 of the 150 cases, the switch occurs at least 0.3 s after ACT.

These observations motivate the identification of a signal time, SigT. The blinker indicates an intention to change lanes, but it does not establish the exact time at which the internal target-gap choice was formed or lateral movement will begin. Therefore, we define SigT in the following way: SigT is the observed switch time for the 23 cases with a delayed switch and ACT for the remaining cases, in which no significant delayed switch is observed. This operational definition identifies a reproducible pre-LCS signal in the trajectory without interpreting it as the exact internal target-gap decision time.

Based on these observations, we conjecture that the tAV lane-change consists of two stages: 
\begin{itemize}
    \item Stage 1: Before LCS, the tAV primarily performs longitudinal preparation, including adjusting its relative position and speed with respect to the target-lane leader and follower. 
    \item Stage 2: After LCS, the vehicle begins the lane-switching process, moving from its original lane to the target lane, during which lateral movement proceeds together with continued longitudinal adjustment until LCE. 
\end{itemize}
	
\begin{figure}[!htbp]
    \centering
    \includegraphics[width=0.8\linewidth]{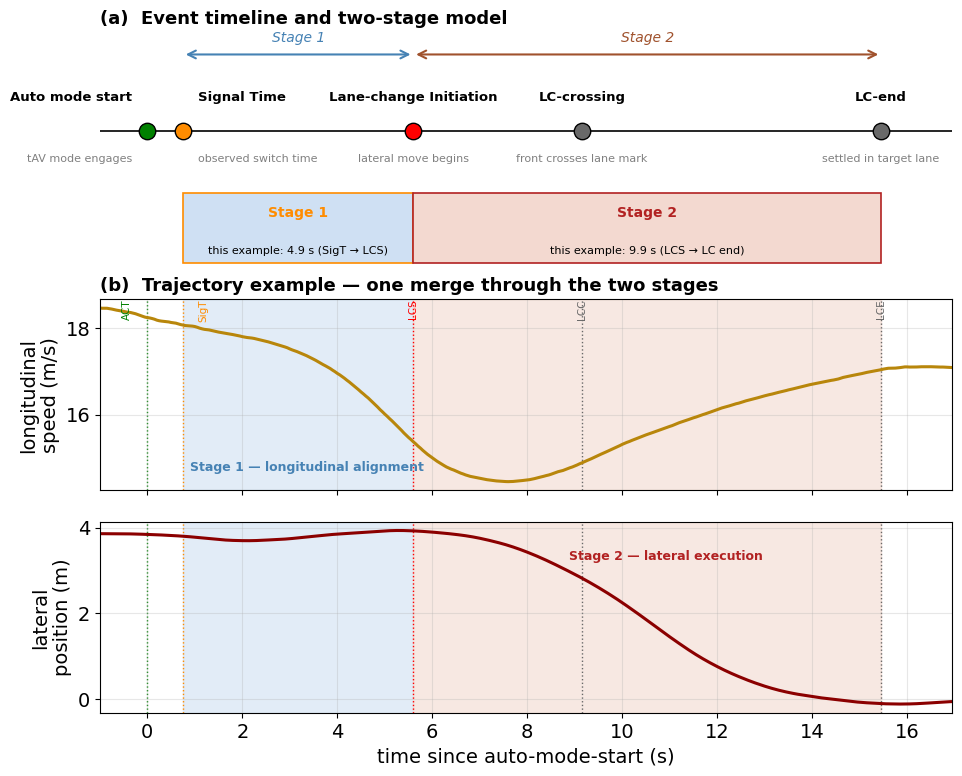}
    \caption{Example illustration of two-stage process with delayed switch}
    \label{fig:overview1}
\end{figure}

Under this conjecture, the target-gap choice is likely substantially resolved before LCS, although its exact internal decision time cannot be observed directly in our data. Figure \ref{fig:overview1} shows the illustration example of the two stages.

\begin{figure}[!htbp]
    \centering
    \includegraphics[width=0.8\linewidth]{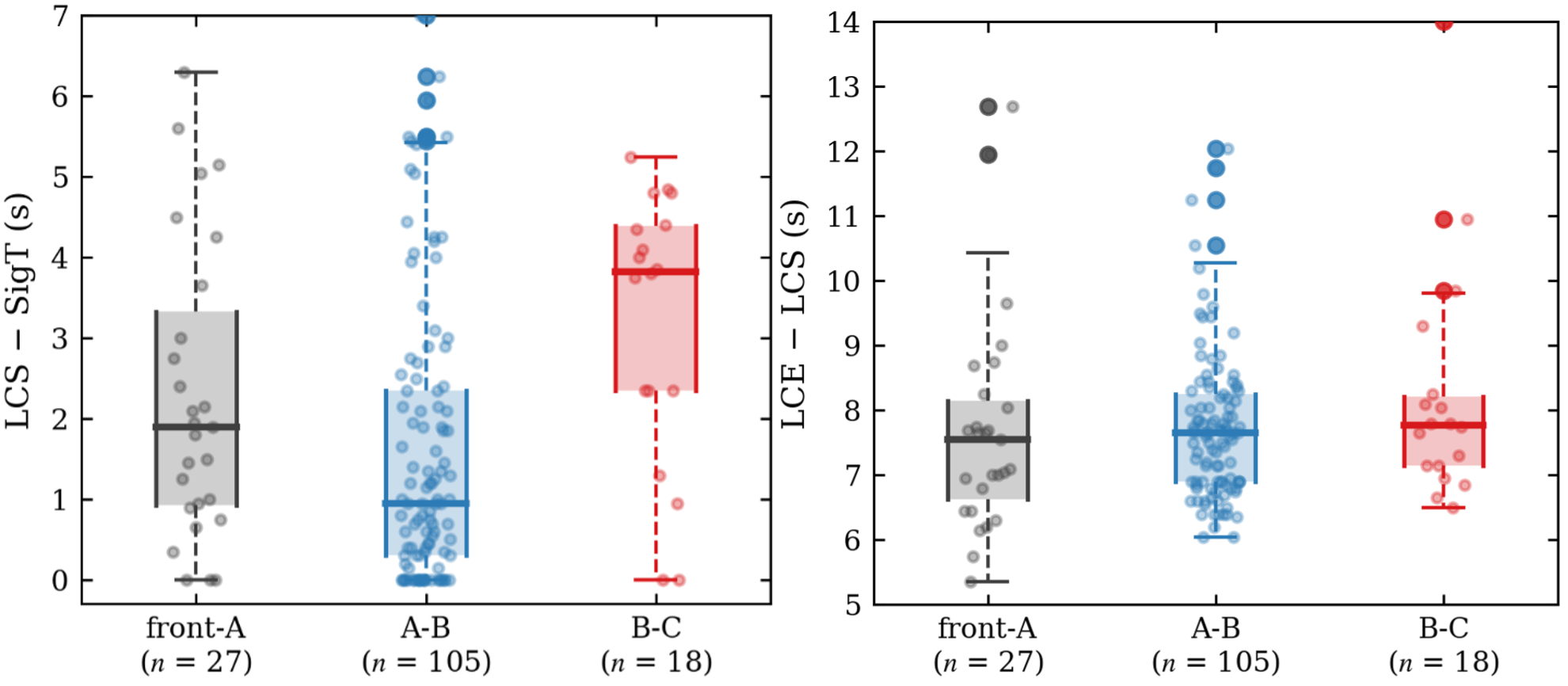}
    \caption{\captitle{Illustration of the two stages.} (a) Pre-LC duration; (b) LCS to LCE. Total sample: $n = 150$.}
    \label{fig:2 stage}
\end{figure}

For the magnitude of the two stages, Figure \ref{fig:2 stage} provides the box plot for the pre-LC duration (from SigT to LCS) and the stage 2 process from LCS to LCE.  One can see that, the pre-LC process can be significant, particularly for those merging into Gap=2, with a median of 3.83 s.  This result suggests that the pre-LC longitudinal preparation can be particularly profound in some conditions.  This further motivates the separation of the two stages. 

The following section tests the Stage 1 implication of this conjecture by examining whether the state at SigT already contains substantial information about the eventual target-gap choice. If so, this would support the inference that the target-gap choice is substantially resolved before LCS.

\section{GAP-CHOICE INFORMATION AT SIGT}\label{sec:framework}

\subsection{Problem Formulation}
This section examines whether the traffic state at SigT contains substantial information about the target gap that X eventually uses.

This gap-choice question is distinct from the one many lane-change decision models pose \cite{Ali2025EmpiricalNeeds}, namely whether X will merge into a given gap now. There, a predicted “yes” bundles two pieces of information—the gap eventually used and whether the LC begins at that time—whereas a “no” identifies neither. Because SigT precedes LCS, we test whether the state observed at SigT already contains information that predicts the eventual target-gap outcome before lateral movement begins.

The model developed here predicts the eventual gap outcome while leaving the exact lateral onset to the next section. It thus separates gap-choice information from lateral-onset timing rather than criticizing timing models. Empirically, in 76\% of merges, X’s center already lies within its final gap at SigT; in the remaining 24\%, it repositions longitudinally to a neighboring gap.
Although the physical alternatives are the front-A, A-B, and B-C gaps, we conjecture that the local gap-choice outcome can be represented relative to the nearest target-lane vehicle N as whether X eventually merges in front of or behind N, coded as

Y = 1, if X eventually merges in front of N; 

Y = 0, if X eventually merges behind N.

The nearest target-lane vehicle is defined as per Equation (\ref{eq:nearest}), where $x_X$ and $x_v$ denote the longitudinal center positions of $X$ and target-lane vehicle $v$, respectively. 
\begin{equation}
\mathrm{N}=\operatorname*{arg\,min}_{v\in\{A,B,C\}}\lvert x_X-x_v\rvert
\label{eq:nearest}
\end{equation}

The two SigT-state features are as per Equation (\ref{eq:xnstate}), where $g_{XN} > 0$ indicates that $X$ is ahead of $N$, and $\Delta v_{XN} > 0$ indicates that $X$ is traveling faster than $N$. This reformulates the multi-gap choice as a binary front/behind prediction problem using $g_{XN}$ and $\Delta v_{XN}$, both evaluated at SigT. Based on this formulation, the following subsection estimates the SigT-based front/behind model and evaluates its predictive performance.

\begin{equation}
g_{X\mathrm N}=\frac{x_X-x_{\mathrm N}}{v_{\mathrm N}},
\qquad
\Delta v_{X\mathrm N}=v_X-v_{\mathrm N}.
\label{eq:xnstate}
\end{equation}

\subsection{SigT-Based Front/Behind Model}
\begin{figure}[!htbp]\centering\includegraphics[width=0.82\linewidth]{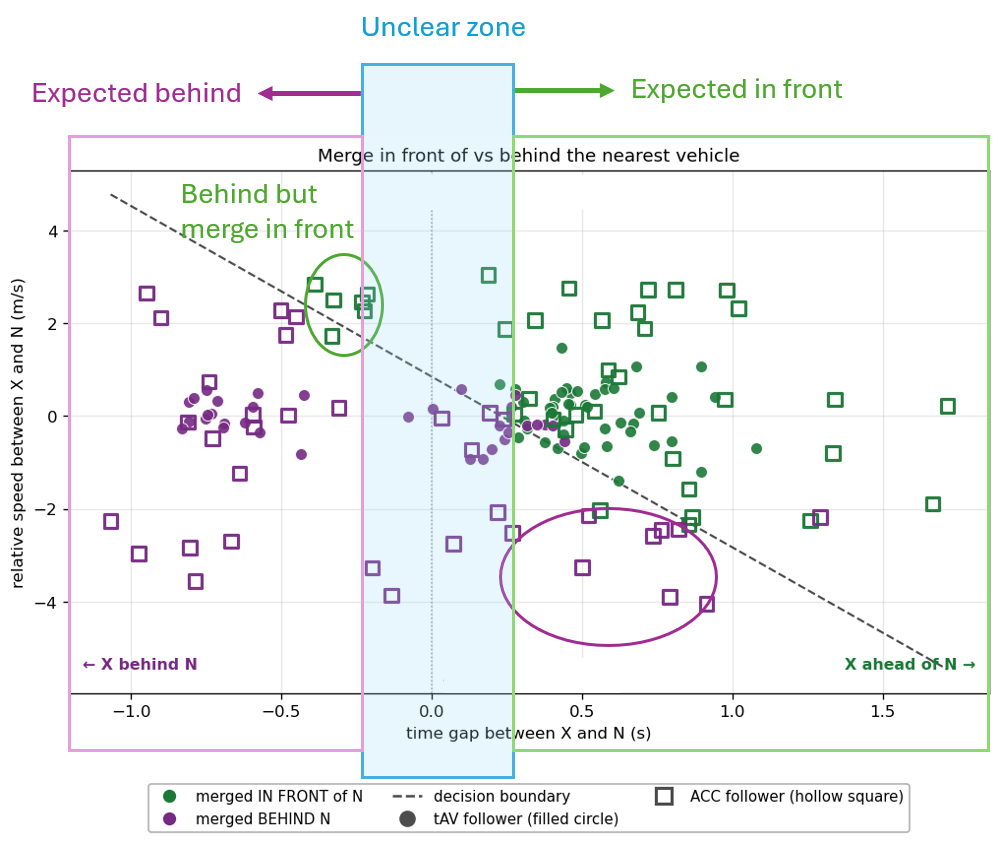}
\caption{\captitle{Front/Behind Gap Selection.} $X$ merges in front of (green) or behind (purple) its nearest vehicle N. Position alone (X-N time gap, $x$-axis) separates the choice at AUC 0.87; adding relative speed ($y$-axis) lifts it to 0.97, the tilt of the fitted boundary (dashed) deciding the near-aligned cases, where $X$ merges ahead when closing on N. Marker shape (filled circle tAV, hollow square ACC) shows the rule holds for both follower types. Shading marks the regime expected from position alone (green: expected in front, $g_{X\mathrm N}>0$; purple: expected behind; blue: the unclear zone near $g_{X\mathrm N}\approx0$, where the bodies still overlap); the ovals highlight unexpected cases where relative speed overrides position (behind N yet merging in front, and ahead of N yet merging behind).}
\label{fig:gapgeom}\end{figure}

The eventual front/behind outcome is strongly associated with X’s position relative to its nearest target-lane vehicle N at SigT: X generally merges on the side of N where it is already located. Because relative position almost separates the outcome, a binary Firth logistic regression—a penalized logistic model that remains finite when a predictor almost perfectly separates the two classes—is fitted using the X-N time gap and relative speed. The fitted model is shown in Equation (\ref{eq:frontbehind}). Both predictors are statistically significant (p < 0.001; Table \ref{tab:firth} and Figure \ref{fig:gapgeom}).

\begin{equation}
\Pr\left(X \text{ eventually merges in front of } N\right)
= \sigma\left(\beta_0 + \beta_1 \tilde{g}_{XN}
+ \beta_2 \Delta\tilde{v}_{XN}\right),
\qquad
\sigma(u) = \frac{1}{1 + e^{-u}}
\label{eq:frontbehind}
\end{equation}

\begin{table}[!htbp]
\caption{\captitle{Front/Behind Selection Model.} Firth logistic regression of Equation~(\ref{eq:frontbehind}); outcome $=1$ if $X$ merges \emph{in front of} its nearest vehicle; standardized predictors. The unpenalized MLE quasi-separates (OR$\approx$498). AUC 0.97; 5-fold accuracy 0.89; leave-one-dataset-out accuracy 0.87 ($n=150$).}
\label{tab:firth}
\centering\small\begin{tabular}{lcccc}\toprule
Term & Firth coef & OR & $z$ & $p$ \\ \midrule
Intercept & $-0.37$ & 0.69 & $-1.1$ & 0.288 \\
X-N time gap & $+5.68$ & 294 & 4.9 & $<$0.001 \\
X-N relative speed & $+3.80$ & 45 & 4.7 & $<$0.001 \\ \bottomrule
\end{tabular}\end{table}

Here, $\tilde{g}_{XN}$ and $\Delta\tilde{v}_{XN}$ are the z-standardized, zero-mean, unit-variance forms of $g_{XN}$ and $\Delta v_{XN}$, respectively. The fitted standardized coefficients are \[
\beta_0 = -0.37, \qquad \beta_1 = 5.68, \qquad \beta_2 = 3.80
\] as per Table \ref{tab:firth}. Because both predictors are standardized, $\beta_1$ and $\beta_2$ are directly comparable, and the larger $\beta_1$ indicates that relative position provides the dominant predictive information.

For the in-sample analysis, the full model attains an AUC of 0.97. Position alone, $g_{XN}$, yields an AUC of 0.87, whereas adding relative speed, $\Delta v_{XN}$, increases the AUC to 0.97. The sign of $g_{XN}$ provides the primary prediction of the eventual outcome: $X$ generally merges in front of $N$ when $g_{XN} > 0$ and behind $N$ when $g_{XN} < 0$. Relative speed mainly helps resolve ambiguous cases, such as those in which $X$ and $N$ are nearly aligned. For example, when $X$ is behind $N$ but very close, a positive $\Delta v_{XN}$ may allow $X$ to move up and merge ahead. This near-deterministic dependence on local geometry contrasts with human gap acceptance, where drivers select among co-existing gaps and can even reject gaps larger than those they eventually take \citep{Marczak2013MergingDevelopment}.

In 5-fold cross-validation, the full model attains an average held-out accuracy of 0.89. The 150 cases are divided into five folds; in each run, the model is fitted using 120 cases and tested on the remaining 30, so every case is tested once.

Leave-one-out cross-validation is used separately to obtain one held-out prediction for each case and construct the confusion matrix, as shown in Table \ref{tab:confusion}. It correctly classifies 135 of the 150 cases, corresponding to an accuracy of 0.90. Of the 15 errors, six realized front cases are predicted behind and nine realized behind cases are predicted front. The leave-one-dataset-out accuracy is 0.87.

\begin{table}[!htbp]
\caption{\captitle{Front/Behind Confusion Matrix.} Leave-one-out cross-validated predictions of the model in Equation~(\ref{eq:frontbehind}); positive $=$ merge \emph{in front of} N. Rows are the realized choice, columns the prediction; 135/150 correct (accuracy 0.90, consistent with the 5-fold 0.89 of Table~\ref{tab:firth}).}
\label{tab:confusion}
\centering\small\begin{tabular}{lccc}\toprule
 & Predicted front & Predicted behind & Total \\ \midrule
Actual front  & 76 (TP) & 6 (FN)  & 82 \\
Actual behind & 9 (FP)  & 59 (TN) & 68 \\ \midrule
Total         & 85      & 65      & 150 \\ \bottomrule
\end{tabular}\end{table}

Overall, the model results show that relative position provides the dominant gap-choice information, while relative speed is particularly useful when X and N are nearly aligned or when X subsequently moves across N. The latter cases are examined next.

\subsection{In-Position and Repositioning Cases}
The repositioning cases further illustrate the role of relative speed. Based on X’s longitudinal position at SigT and the eventual target gap, the cases are classified as in-position or repositioning. An in-position case is one in which X is already between the leader and follower of its eventual target gap at SigT, as per Equation (\ref{eq:inposition}).
\begin{equation}
x_{\mathrm{foll}} \leq x_X \leq x_{\mathrm{lead}}
\label{eq:inposition}
\end{equation}

A repositioning case is one in which $X$ occupies a different gap region at SigT and subsequently moves longitudinally across a target-lane vehicle into a neighboring gap.

Among the 150 cases, 114 are in-position and 36 are repositioning cases. For the 36 repositioning cases, the absolute longitudinal separation between $X$ and the nearest target-lane vehicle $N$, $\left|x_X - x_N\right|$, has a median of 4.8~m at SigT, and 75\% of the cases have $\left|x_X - x_N\right| \leq 8$~m. Thirty cases are dropbacks: 15 move from ahead of $A$ into the $A$--$B$ gap, whereas the other 15 move from ahead of $B$ into the $B$--$C$ gap. The remaining six cases are move-ups: three move from the $A$--$B$ gap to the front-$A$ gap, and three move from the $B$--$C$ gap to the $A$--$B$ gap.

\begin{figure}[!htbp]\centering\includegraphics[width=\linewidth]{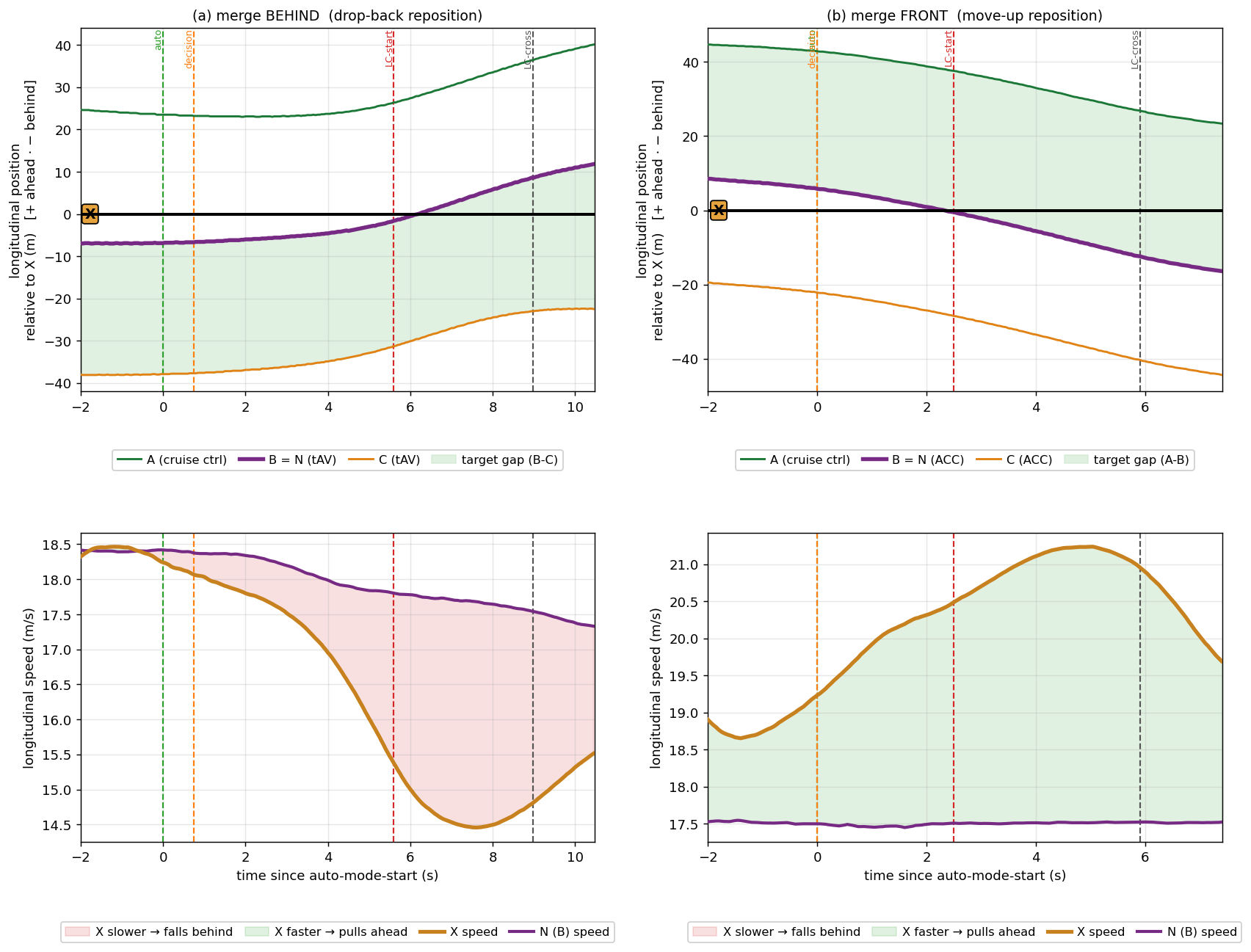}
\caption{\captitle{Reposition Examples.} In a reposition, $X$ moves longitudinally across its nearest vehicle N to settle inside the target gap (shaded); each panel plots target-lane vehicles' longitudinal position relative to $X$ (bold 0 line, $+$ ahead / $-$ behind), so N crossing 0 marks the alignment. (a) \emph{Merge behind} (drop-back): $X$ decelerates and settles behind N; (b) \emph{merge in front} (move-up): $X$ accelerates and settles ahead of N. Legends give vehicle types (A is the cruise-control lead; B and C are tAV followers in (a), ACC in (b)).}
\label{fig:repos}\end{figure}

The results are detailed in Table \ref{tab:commitstate}. Figure \ref{fig:repos} illustrates a representative dropback and move-up.

\begin{table}[!htbp]
\caption{\captitle{Gap-Decision States by Gap.}}
\label{tab:commitstate}
\centering\small\begin{tabular}{llcccc}\toprule
Class & Direction & front-A & A-B & B-C & Total \\ \midrule
\textbf{in-position} & n/a & 24 & 87 & 3 & \textbf{114} \\
\textbf{reposition} & drop-back & 0 & 15 & 15 & 30 \\
\textbf{reposition} & move-up & 3 & 3 & 0 & 6 \\ \bottomrule
\end{tabular}\end{table}

The direction of repositioning is consistent with $X$'s relative speed to the vehicle it crosses. In the 30 dropback cases, $X$ is generally slower than the crossed vehicle, with a median $\Delta v_{XN}$ of $-0.6$~m/s. In the six move-up cases, $X$ is faster than the crossed vehicle, with a median $\Delta v_{XN}$ of $2.5$~m/s. Thus, relative speed helps explain whether $X$ moves forward or backward across the nearby vehicle when its gap occupancy at SigT differs from its eventual target gap.

These repositioning results explain how the state at SigT can predict an eventual gap that differs from the gap region occupied by $X$ at that time. The next subsection examines how far in advance of LCS this information is observable.

\subsection{Lead Time of Gap-Choice Information}
\begin{figure}[!htbp]\centering\includegraphics[width=0.6\linewidth]{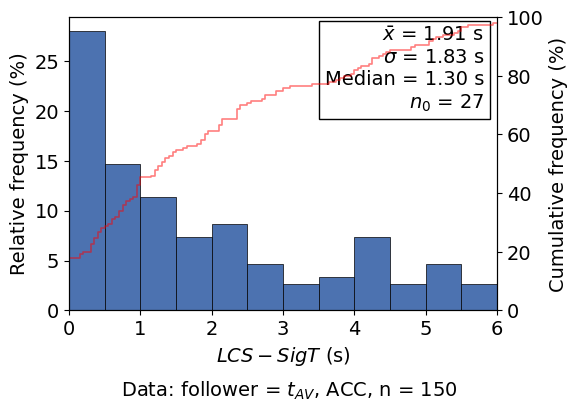}
\caption{\captitle{Gap-Choice Lead Time.} Distribution of $\mathrm{LC\text{-}start}-\mathrm{GT}$ across the 150 merges, the interval by which the front/behind gap choice is fixed before lateral motion begins: 38.7\% exceed 2\,s and 24\% exceed 3\,s (median 1.3\,s), so the gap-selection signal is systematically available ahead of LC-start.}
\label{fig:leadtime}\end{figure}

Since the state at SigT strongly predicts the eventual front/behind merging outcome, the results suggest that substantial gap-choice information is probably observable before lateral movement begins, identified as lane-change start, LCS. The corresponding lead interval is $\mathrm{LCS} - \mathrm{SigT}$ as shown in Figure~\ref{fig:leadtime}.

In these data, the lead interval is often substantial: 38.7\% of cases have $\mathrm{LCS} - \mathrm{SigT} > 2$~s, 24\% exceed 3~s, and the median is 1.3~s.

We caution that these fractions pertain to the controlled experimental scenarios and should not be interpreted as naturalistic rates. The results show that substantial gap-choice information is systematically observable at SigT before LCS. The observed lead interval likely represents a lower bound on how early such information could have been available if automation had been activated earlier, given that SigT equals ACT in most cases in our dataset.

The lead-time results establish when the information is observable. The final subsection considers why the front/behind representation is meaningful relative to conventional current-gap formulations.

\subsection{Implications Relative to Conventional LC Decision Models}
To further understand the implications of the model, consider the typical representation used in conventional LC decision models, such as gap-acceptance and game-theoretic models \cite{zheng2017gapchoice, TALEBPOUR2015, ALI2019}. These models commonly evaluate whether a vehicle will merge into the current candidate gap it faces. This candidate gap typically has positive lead and lag bumper-to-bumper clearances. Namely, X is fully within the gap between a target-lane leader and follower when condition in Equation (\ref{eq:withineq}) satisfies, where x denotes the longitudinal center position and l is the vehicle length, assumed to be the same for all vehicles.
\begin{equation}
x_{\mathrm{foll}} + l \leq x_X \leq x_{\mathrm{lead}} - l
\label{eq:withineq}
\end{equation}

Outside the gap, there are also unclear zones in which X overlaps longitudinally with the leader or follower; that is as per Equation (\ref{eq:outsidegap1}) or (\ref{eq:outsidegap2}).
\begin{equation}
x_{\mathrm{lead}} - l < x_X < x_{\mathrm{lead}} + l
\label{eq:outsidegap1}
\end{equation}

\begin{equation}
x_{\mathrm{foll}} - l < x_X < x_{\mathrm{foll}} + l
\label{eq:outsidegap2}
\end{equation}

In these unclear zones, the bumper-to-bumper clearance between X and the overlapping vehicle is negative.

For cases in the unclear zone, conventional LC decision models may not clearly define the current candidate gap because the bumper-to-bumper clearance is negative. However, our results indicate that prediction of the eventual gap choice remains possible for cases in this zone; one can see that the front and behind outcomes remain well separated by the decision boundary within the unclear zone in Figure \ref{fig:gapgeom}.
           
For cases in the within-a-gap zone, conventional LC decision models represent a vehicle as evaluating the current candidate gap and deciding whether to merge into it or reject it. If the gap is rejected, the vehicle applies the same evaluation process to the next gap it encounters. Our results show, however, that X may be evaluating both the current gap and a neighboring gap and may move up or drop back, which may appear “surprising” in the conventional LC decision context. For example, the cases in the purple oval may appear surprising: X is significantly ahead of N at SigT—the cases are located at the far right—but eventually merges behind N. Similarly, the cases in the green oval may contradict the baseline expectation because X is behind N at SigT but eventually merges ahead.

Taken together, the results show that the SigT-based front/behind formulation remains informative across three settings: when X is clearly within a candidate gap and ultimately merges into that gap, when X repositions into a neighboring gap, and when X is in an ambiguous zone where it longitudinally overlaps a target-lane vehicle. The formulation therefore provides a common representation of the eventual target-gap outcome across cases that are straightforward, surprising, or geometrically unclear under a conventional current-gap perspective. More broadly, the strong predictive performance at SigT supports the conjecture that substantial target-gap information is already present before LCS.

\section{LONGITUDINAL PREPARATION FROM SigT TO LCS}

\subsection{Motivation}
Section 4 showed that the traffic state at SigT contains substantial information about the eventual target-gap outcome. The remaining question is how X progresses from SigT to the onset of lateral movement at LCS.

We conjecture that LCS may correspond to a recognizable lateral-onset state. To examine this conjecture, the analysis distinguishes the two groups identified in Section 4: the 114 in-position cases, in which X is already within the region of its eventual target gap at SigT, and the 36 repositioning cases, in which X occupies a neighboring gap region at SigT and subsequently moves longitudinally across a target-lane vehicle. The analysis examines whether these groups follow different longitudinal processes from SigT to LCS and whether they reach similar or different states when lateral movement begins.

\subsection{In-Position Cases}
For the in-position cases, $X$ is already located between the leader and follower of its eventual target gap at SigT. Initial analyses considered several candidate variables describing the longitudinal state from SigT to LCS. These analyses showed that the most relevant variables are the safety margins on the two sides of the selected gap.

Specifically, we adopt a safety surrogate measure proposed in the Overview paper \cite{sharma2026controlled}, after emergency spacing (AES). This variable measures whether the available longitudinal spacing between a leader and follower is sufficient to avoid a rear-end collision under a hypothetical emergency-braking scenario. For a leader $l_d$ and follower $f$, AES at time $t$ is defined as per Equation (\ref{eq:AES}).

\begin{equation}
\mathrm{AES}_{l_d f}(t)
=
\left|x_{l_d}(t) - x_f(t) - l - b_0\right|
-
\left[
v_f(t)\tau_e
+ \frac{1}{2}a_f(t)\tau_e^2
+ \frac{\left(v_f(t) + a_f(t)\tau_e\right)^2}{2b_f}
- \frac{v_{l_d}(t)^2}{2b_{l_d}}
\right]
\label{eq:AES}
\end{equation}

where $x_{l_d}$ and $x_f$ are the longitudinal center positions of the leader and follower; $v_{l_d}$ and $v_f$ are their speeds; $a_f$ is the follower acceleration; $l$ is the vehicle length; $b_0$ is an additional spacing buffer; $\tau_e$ is the follower response time; and $b_{l_d}$ and $b_f$ are the emergency-deceleration magnitudes of the leader and follower, respectively. The formulation assumes that the leader begins emergency braking immediately, while the follower continues under its current acceleration during the response interval and then brakes. A negative AES indicates insufficient projected spacing, whereas a larger AES indicates a greater safety margin.

For each in-position case, $\mathrm{AES}_{\mathrm{lead}}$ denotes the AES for the selected-gap leader--$X$ pair, whereas $\mathrm{AES}_{\mathrm{lag}}$ denotes the AES for the $X$--selected-gap follower pair. The binding AES is defined as per Equation (\ref{eq:bindingAES}).
\begin{equation}
\mathrm{AES}_{\mathrm{bind}}
=
\min\left(
\mathrm{AES}_{\mathrm{lead}},
\mathrm{AES}_{\mathrm{lag}}
\right)
\label{eq:bindingAES}
\end{equation}

The binding AES represents the more restrictive of the two longitudinal safety margins.

\begin{figure}[!htbp]\centering\includegraphics[width=0.80\linewidth]{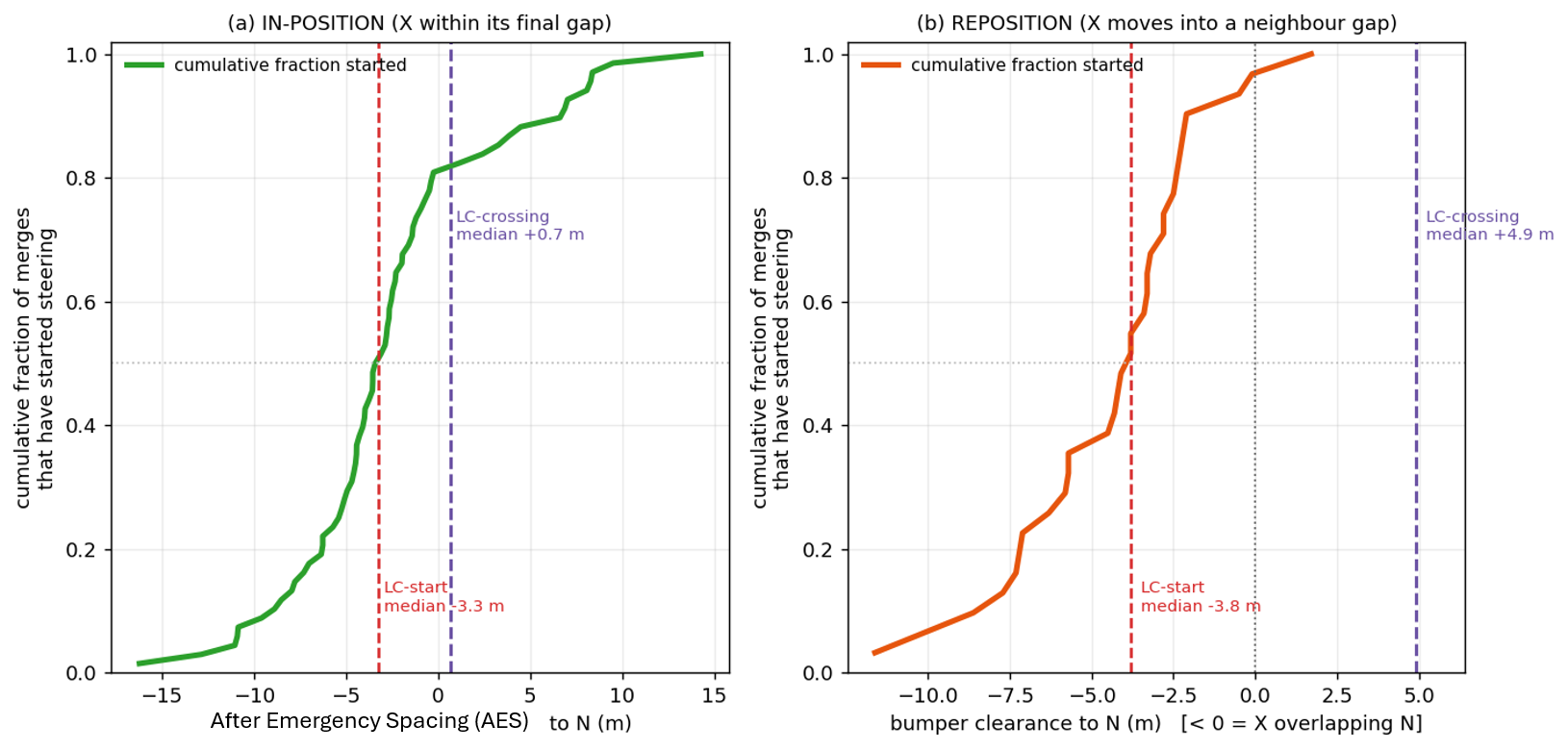}
\caption{\captitle{The Lateral-Onset Threshold.} Empirical CDF of the margin to the binding vehicle at LC-start (the steep part is the threshold). (a) In-position: merges begin as the AES recovers past $\approx-3$\,m, half steering at a small negative deficit rather than a positive buffer. (b) Reposition: merges begin as bumper clearance to N crosses $\approx-4$\,m ($X$ still slightly overlapping N). One rule on opposite sides: $X$ steers at a near-zero deficit (red median at LC-start) and recovers to a cleared margin by LC-crossing (purple median, $+0.7$\,m in-position, $+4.9$\,m reposition), completing the longitudinal adjustment during the lateral move.}
\label{fig:onset}\end{figure}

Figure \ref{fig:onset}(a) shows the cumulative fraction of cases in which lateral movement has begun as a function of the binding AES. The fraction rises most strongly as the binding AES approaches approximately $-3$~m. At LCS, the median binding AES is also approximately $-3$~m. Thus, lateral movement often begins while the binding AES remains slightly negative rather than after a positive margin has been reached. The distribution nevertheless remains dispersed, indicating that the observed value is not a sharp threshold.

An exploratory prediction analysis provides supporting evidence on the association between the observed traffic state and LCS. Binding AES is the strongest individual indicator for the in-position cases, with an AUC of approximately 0.71. Specifications using additional instantaneous variables attain AUC values of approximately 0.74–0.76. These results indicate moderate, rather than strong, discrimination of the exact lateral-onset time.

\subsection{Repositioning Cases}
For the repositioning cases, X must move longitudinally across a target-lane vehicle before entering the region of its eventual target gap. Initial analyses considered several candidate variables describing the longitudinal state from SigT to LCS. These analyses showed that the most relevant variable is the signed bumper-to-bumper clearance to the crossed vehicle. We therefore examine this clearance, whether longitudinal overlap remains at LCS, and X’s position relative to the boundary of the selected gap.

For each repositioning case, let N denote the target-lane vehicle that X crosses. The signed bumper-to-bumper clearance between X and N is defined as per Equation (\ref{eq:bumperclearance}).
\begin{equation}
c_{XN}(t) = \left|x_X(t) - x_N(t)\right| - l
\label{eq:bumperclearance}
\end{equation}

where $x_X$ and $x_N$ are the longitudinal center positions of $X$ and $N$, respectively, and $l$ is the vehicle length, assumed to be the same for both vehicles. A positive $c_{XN}$ indicates that $X$ has longitudinally cleared $N$, whereas a negative value indicates that the two vehicles still overlap longitudinally.

Figure \ref{fig:onset}(b) shows the cumulative fraction of cases in which lateral movement has begun as a function of $c_{XN}$. The fraction rises most strongly as the clearance approaches approximately $-4$~m. At LCS, the median clearance is also approximately $-4$~m. Thus, lateral movement often begins while $X$ still overlaps longitudinally with the crossed vehicle. The distribution nevertheless remains dispersed, indicating that the observed value is not a sharp threshold.

An exploratory prediction analysis provides supporting evidence on the association between the observed traffic state and LCS. The signed bumper-to-bumper clearance is the strongest individual indicator for the repositioning cases, with an AUC of approximately 0.84. Specifications using additional instantaneous variables attain AUC values of approximately 0.84--0.87. These results indicate a stronger association with LCS than that observed for the in-position cases, although the repositioning group contains only 36 cases.

\subsection{Comparison, Summary, and Future Research}
Taken together, the results provide some evidence that the in-position and repositioning cases follow different longitudinal pathways from SigT to LCS. The in-position cases are characterized by changes in the safety margins within the selected gap, whereas the repositioning cases are characterized by longitudinal movement relative to the crossed vehicle.

However, the observed relationships are not strong enough to establish a single clear lateral-onset condition for either group. The findings therefore indicate group-dependent patterns in the progression toward LCS, while also showing that the observed longitudinal measures do not fully characterize when lateral movement begins. Future research should examine additional state variables, the temporal evolution of the observed state before LCS, and larger samples of repositioning cases to determine whether a more consistent lateral-onset pattern can be identified.

\section{Conclusions and Future Work}\label{sec:conclusion}
Empirical understanding of how production tAVs develop and execute lane-change decisions remains limited. To address this gap, this study analyzes 150 controlled mandatory lane changes performed by a production tAV in the NC-tALC experiments. The analysis focuses on two process-level questions: whether information about the eventual target gap is observable before lateral movement begins, and how the tAV progresses longitudinally from that pre-LCS state to the onset of lateral movement. 

To support this analysis, the signal time (SigT), which occurs prior to LC start, is used as an operationally identifiable pre-LCS reference point.  Our analysis shows that the traffic state at SigT contains substantial information about the tAV’s eventual target-gap choice, providing meaningful lead time before lateral movement begins. The proposed formulation well predicts whether the tAV remains with the gap it occupies at SigT or repositions to a neighboring gap by moving forward or dropping back. It also applies when the tAV longitudinally overlaps a target-lane vehicle and the current-gap geometry is ambiguous (such cases were considered invalid in critical gap theories).  These findings support a two-stage conjecture for the observable LC process: longitudinal preparation from SigT to LCS, followed by lateral maneuver execution from LCS to LCE. The study further conjectures that a target-gap decision-making process may precede longitudinal preparation, although this cannot be evaluated directly with the available data, which calls for future research.  

This study also provides preliminary evidence that in-position and repositioning cases follow different longitudinal pathways from SigT to LCS, with the former associated with a surrogate safety measure and the latter associated with a bumper clearance.  These results motivate further research on the longitudinal preparation process. 

We caution that, the results here are based on controlled experiments covering a limited set of scenarios, rather than naturalistic driving. They are intended mainly to reveal recurrent behaviors, not to provide precise naturalistic estimates.  For example, in many cases, LCS occurred immediately after ACT. This limits how early SigT can be observed. Earlier automation activation will likely shift SigT earlier and help determine how far in advance target-gap information can be inferred. Also, repositioning cases represent only a small fraction of the sample. This reflects the experimental design and does not indicate their frequency in naturalistic traffic.  Thus, our findings should not be generalized directly to other tAV systems, software versions, roadway environments, or traffic conditions. The analysis of how the tAV progresses from LCS to LCE remains preliminary and requires more thorough investigation.

Nevertheless, the results suggest that the LC behaviors of tAV may differ from those of human-driven vehicles and that motivate additional empirical studies (either controlled experiments or extensive naturalistic studies), including extending the observation period further before LCS to better understand LC decision making, testing a broader range of initial conditions to generate more diverse gap choices, including repositioning, and examining additional tAV systems.

\section*{ACKNOWLEDGMENTS}

This material is based upon work supported by the National Science Foundation under Award No.~2401555.

\section{Author Contribution Statement}
\textbf{Zeyu Mu:} Conceptualization, Methodology, Investigation, Formal Analysis, Writing--review \& editing.
\textbf{Danjue Chen:} Conceptualization, Funding acquisition, Methodology, Investigation, Project administration, Resources, Supervision, Writing--review \& editing.
\textbf{Abhinav Sharma:} Visualization, Writing--review \& editing.
\textbf{George F. List:} Conceptualization, Supervision, Methodology, Writing--review \& editing.

\newpage
\bibliographystyle{trb_chicago.bst}
\bibliography{merge_refs, trb_template, luo_literature}

@article{Kim2008,
author = {Kim, Jin-Tae and Kim, Joonhyon and Chang, Myungsoon},
title = {Lane-changing gap acceptance model for freeway merging in simulation},
journal = {Canadian Journal of Civil Engineering},
volume = {35},
number = {3},
pages = {301-311},
year = {2008},
doi = {10.1139/L07-119},

URL = {https://doi.org/10.1139/L07-119},
eprint = {https://doi.org/10.1139/L07-119}
}

@article{Hess2020,
title = {Modelling lane changing behaviour in approaches to roadworks: Contrasting and combining driving simulator data with stated choice data},
journal = {Transportation Research Part C: Emerging Technologies},
volume = {112},
pages = {282-294},
year = {2020},
issn = {0968-090X},
doi = {https://doi.org/10.1016/j.trc.2019.12.003},
url = {https://www.sciencedirect.com/science/article/pii/S0968090X19306874},
author = {Stephane Hess and Charisma F. Choudhury and Michiel C.J. Bliemer and Daryl Hibberd}
}

@article{TOLEDO2007,
title = {Integrated driving behavior modeling},
journal = {Transportation Research Part C: Emerging Technologies},
volume = {15},
number = {2},
pages = {96-112},
year = {2007},
issn = {0968-090X},
doi = {https://doi.org/10.1016/j.trc.2007.02.002},
url = {https://www.sciencedirect.com/science/article/pii/S0968090X07000046},
author = {Tomer Toledo and Haris N. Koutsopoulos and Moshe Ben-Akiva}
}

@article{KITA1999,
title = {A merging–giveway interaction model of cars in a merging section: a game theoretic analysis},
journal = {Transportation Research Part A: Policy and Practice},
volume = {33},
number = {3},
pages = {305-312},
year = {1999},
issn = {0965-8564},
doi = {https://doi.org/10.1016/S0965-8564(98)00039-1},
url = {https://www.sciencedirect.com/science/article/pii/S0965856498000391},
author = {Hideyuki Kita}
}

@article{TALEBPOUR2015,
title = {Modeling lane-changing behavior in a connected environment: A game theory approach},
journal = {Transportation Research Part C: Emerging Technologies},
volume = {59},
pages = {216-232},
year = {2015},
note = {Special Issue on International Symposium on Transportation and Traffic Theory},
issn = {0968-090X},
doi = {https://doi.org/10.1016/j.trc.2015.07.007},
url = {https://www.sciencedirect.com/science/article/pii/S0968090X15002478},
author = {Alireza Talebpour and Hani S. Mahmassani and Samer H. Hamdar}
}

@article{PENG2015,
title = {Multi-parameter prediction of drivers' lane-changing behaviour with neural network model},
journal = {Applied Ergonomics},
volume = {50},
pages = {207-217},
year = {2015},
issn = {0003-6870},
doi = {https://doi.org/10.1016/j.apergo.2015.03.017},
url = {https://www.sciencedirect.com/science/article/pii/S0003687015000502},
author = {Jinshuan Peng and Yingshi Guo and Rui Fu and Wei Yuan and Chang Wang}
}

@article{Wirthmuller2021,
  author={Wirthmüller, Florian and Klimke, Marvin and Schlechtriemen, Julian and Hipp, Jochen and Reichert, Manfred},
  journal={IEEE Robotics and Automation Letters}, 
  title={Predicting the Time Until a Vehicle Changes the Lane Using LSTM-Based Recurrent Neural Networks}, 
  year={2021},
  volume={6},
  number={2},
  pages={2357-2364},
  doi={10.1109/LRA.2021.3058930}}

@ARTICLE{Mozaffari2022,
  author={Mozaffari, Sajjad and Arnold, Eduardo and Dianati, Mehrdad and Fallah, Saber},
  journal={IEEE Transactions on Intelligent Vehicles}, 
  title={Early Lane Change Prediction for Automated Driving Systems Using Multi-Task Attention-Based Convolutional Neural Networks}, 
  year={2022},
  volume={7},
  number={3},
  pages={758-770},
  doi={10.1109/TIV.2022.3161785}}

@article{Bakhit2017,
  author  = {Bakhit, Peter R. and Osman, Osama A. and Ishak, Sherif},
  title   = {Detecting Imminent Lane Change Maneuvers in Connected Vehicle Environments},
  journal = {Transportation Research Record: Journal of the Transportation Research Board},
  year    = {2017},
  volume  = {2645},
  number  = {1},
  pages   = {168--175},
  doi     = {10.3141/2645-18},
  url     = {https://doi.org/10.3141/2645-18}
}

@article{zheng2017gapchoice,
  author  = {Zheng, Zuduo and Sun, Zhongxiang and Xie, Dong-Fan},
  title   = {Modeling Merge Gap Choice Behavior with Latent Choice Sets},
  journal = {Journal of Transportation Engineering, Part A: Systems},
  volume  = {143},
  number  = {9},
  year    = {2017},
}

@article{sharma2026controlled,
  title   = {Controlled Experiments on Lane Changing by Transitional Autonomous Vehicle: Dataset and Behavioral Insights},
  author  = {Sharma, Abhinav and Al Hasan, Md Abdullah and Chen, Danjue and List, George F.},
  journal = {arXiv preprint arXiv:2607.27085},
  year    = {2026},
  doi     = {10.48550/arXiv.2607.27085},
  url     = {https://doi.org/10.48550/arXiv.2607.27085}
}

@article{ALI2023-2issues,
title = {Calibrating lane-changing models: Two data-related issues and a general method to extract appropriate data},
journal = {Transportation Research Part C: Emerging Technologies},
volume = {152},
pages = {104182},
year = {2023},
issn = {0968-090X},
doi = {https://doi.org/10.1016/j.trc.2023.104182},
url = {https://www.sciencedirect.com/science/article/pii/S0968090X23001717},
author = {Yasir Ali and Zuduo Zheng and Michiel C.J. Bliemer}
}

@article{Wan2017,
author = {Xia Wan  and Peter J. Jin  and Haiyan Gu  and Xiaoxuan Chen  and Bin Ran },
title = {Modeling Freeway Merging in a Weaving Section as a Sequential Decision-Making Process},
journal = {Journal of Transportation Engineering, Part A: Systems},
volume = {143},
number = {5},
pages = {05017002},
year = {2017},
doi = {10.1061/JTEPBS.0000048},

URL = {https://ascelibrary.org/doi/abs/10.1061/JTEPBS.0000048},
eprint = {https://ascelibrary.org/doi/pdf/10.1061/JTEPBS.0000048}
}

@article{Chu2017,
author = {Tien Dung Chu  and Tomio Miwa  and Takayuki Morikawa },
title = {Discrete Choice Models for Gap Acceptance at Urban Expressway Merge Sections Considering Safety, Road Geometry, and Traffic Conditions},
journal = {Journal of Transportation Engineering, Part A: Systems},
volume = {143},
number = {7},
pages = {04017025},
year = {2017},
doi = {10.1061/JTEPBS.0000053},

URL = {https://ascelibrary.org/doi/abs/10.1061/JTEPBS.0000053},
eprint = {https://ascelibrary.org/doi/pdf/10.1061/JTEPBS.0000053}
}

@article{Choudhury2009,
author = {Charisma F. Choudhury and Varun Ramanujam and Moshe E. Ben-Akiva},
title ={Modeling Acceleration Decisions for Freeway Merges},

journal = {Transportation Research Record},
volume = {2124},
number = {1},
pages = {45-57},
year = {2009},
doi = {10.3141/2124-05},

URL = { 
    
        https://doi.org/10.3141/2124-05
    
    

},
eprint = { 
    
        https://doi.org/10.3141/2124-05
    
    

}
}

@article{ALI2021,
title = {CLACD: A complete LAne-Changing decision modeling framework for the connected and traditional environments},
journal = {Transportation Research Part C: Emerging Technologies},
volume = {128},
pages = {103162},
year = {2021},
issn = {0968-090X},
doi = {https://doi.org/10.1016/j.trc.2021.103162},
url = {https://www.sciencedirect.com/science/article/pii/S0968090X21001807},
author = {Yasir Ali and Zuduo Zheng and Md. Mazharul Haque and Mehmet Yildirimoglu and Simon Washington}
}

@article{ALI2019,
title = {A game theory-based approach for modelling mandatory lane-changing behaviour in a connected environment},
journal = {Transportation Research Part C: Emerging Technologies},
volume = {106},
pages = {220-242},
year = {2019},
issn = {0968-090X},
doi = {https://doi.org/10.1016/j.trc.2019.07.011},
url = {https://www.sciencedirect.com/science/article/pii/S0968090X19302244},
author = {Yasir Ali and Zuduo Zheng and Md. Mazharul Haque and Meng Wang}
}

@misc{FordMotorCompany2025BlueCruise,
    title = {{BlueCruise}},
    year = {2025},
    author = {{Ford Motor Company}},
    url = {https://www.ford.com/technology/bluecruise/},
    howpublished = {https://www.ford.com/technology/bluecruise/}
}

@misc{Ali2025EmpiricalNeeds,
    title = {{Empirical research on car-following and lane-changing: Recent developments, emerging vehicle technologies’ impact, and future research needs}},
    year = {2025},
    booktitle = {Transportation Research Interdisciplinary Perspectives},
    author = {Ali, Yasir and Sharma, Anshuman and Zheng, Zuduo},
    month = {5},
    volume = {31},
    publisher = {Elsevier Ltd},
    doi = {10.1016/j.trip.2025.101368},
    issn = {25901982}
}

@misc{GeneralMotors2025HowCruise,
    title = {{How to Use Super Cruise}},
    year = {2025},
    author = {{General Motors}},
    url = {https://www.chevrolet.com/support/vehicle/driving-safety/driver-assistance/how-to-use-super-cruise},
    howpublished = {https://www.chevrolet.com/support/vehicle/driving-safety/driver-assistance/how-to-use-super-cruise}
}

@article{Ammourah2024IntroductionExtraction,
    title = {{Introduction to the Third Generation Simulation Dataset: Data Collection and Trajectory Extraction}},
    year = {2024},
    journal = {Transportation Research Record: Journal of the Transportation Research Board},
    author = {Ammourah, Rami and Beigi, Pedram and Fan, Bingyi and Hamdar, Samer H. and Hourdos, John and Hsiao, Chun-Chien and James, Rachel and Khajeh-Hosseini, Mohammdreza and Mahmassani, Hani S. and Monzer, Dana and Radvand, Tina and Talebpour, Alireza and Yousefi, Mahdi and Zhang, Yanlin},
    month = {7},
    doi = {10.1177/03611981241257257},
    issn = {0361-1981}
}

@article{Marczak2013MergingDevelopment,
    title = {{Merging behaviour: Empirical comparison between two sites and new theory development}},
    year = {2013},
    journal = {Transportation Research Part C: Emerging Technologies},
    author = {Marczak, Florian and Daamen, Winnie and Buisson, Christine},
    pages = {530--546},
    volume = {36},
    publisher = {Elsevier Ltd},
    doi = {10.1016/j.trc.2013.07.007},
    issn = {0968090X}
}

@misc{TeslaInc.2025ModelSupervised,
    title = {{Model 3 Owner's Manual: Full Self-Driving (Supervised)}},
    year = {2025},
    author = {{Tesla Inc.}},
    url = {https://www.tesla.com/ownersmanual/model3/en_us/GUID-2CB60804-9CEA-4F4B-8B04-09B991368DC5.html},
    howpublished = {https://www.tesla.com/ownersmanual/model3/en{\_}us/GUID-2CB60804-9CEA-4F4B-8B04-09B991368DC5.html}
}

@techreport{Mattas2025SafetyExperiments,
    title = {{Safety Perspective on Assisted Lane Changes: Insights from Open-Road, Live-Traffic Experiments}},
    year = {2025},
    author = {Mattas, Konstantinos and Vass, Sandor and Zach{\'{a}}r, Gergely and Ji, Junyi and Gloudemans, Derek and Maggi, Davide and Kriston, Akos and Brahmi, Mohamed and Galassi, Maria Christina and Work, Daniel B and Ciuffo, Biagio}
}

@article{ToledoKatz2009,
  author  = {Toledo, Tomer and Katz, Romina},
  title   = {State Dependence in Lane-Changing Models},
  journal = {Transportation Research Record: Journal of the Transportation Research Board},
  volume  = {2124},
  number  = {1},
  pages   = {81--88},
  year    = {2009},
  doi     = {10.3141/2124-08},
  url     = {https://doi.org/10.3141/2124-08}
}
\end{document}